# A specialized reasoning large language model for accelerating rare disease diagnosis: a randomized AI physician-assistance trial

**Haichao Chen MD[1,2*], Songchi Zhou PhD[3*], Zhengyun Zhao PhD[3*], Shikai Hu MD[4], Xianghong Jin MD[5], Hongwei Ji[1], Li He[6], Shuli Li[6], Yiming Qin PhD[1], Xin Tan PhD[7], Runfeng Shi MD[8], Yih Chung Tham PhD[9,10,11], Jiaye Zhu[12], Ye Li MD[8], Ye Jin PhD[5], Longhao Cao[13], Dawei Li[13], Honghan Wu[14], Hongqiu Gu[15], Guanqiao Li[16], Tudor Groza PhD[17,18], Chunying Li MD[6], Dian Zeng[1], Weihong Yu[2], Gareth Baynam PhD[19], Saumya Shekhar Jamuar PhD[17,20], Min Shen MD[5], Shuyang Zhang MD[5**], Bin Sheng PhD[21,22**], Sheng Yu PhD[3**], Tien Yin Wong PhD[1,10**]**

[1]Tsinghua Medicine, Tsinghua University, Beijing, China.

[2]Department of Ophthalmology, Peking Union Medical College Hospital, Chinese Academy of Medical Sciences & Peking Union Medical College, Beijing, China.

[3]Department of Statistics and Data Science, Tsinghua University, Beijing, China.

[4]Department of Rheumatology and Clinical Immunology, Peking Union Medical College Hospital, Chinese Academy of Medical Sciences & Peking Union Medical College, Beijing, China.

[5]Department of Rare Diseases, Peking Union Medical College Hospital, Chinese Academy of Medical Sciences & Peking Union Medical College, Beijing, China.

[6]Department of Dermatology, Xijing Hospital, Air Force Medical University, Xi'an, Shanxi, China.

[7]School of Computer Science and Technology, East China Normal University (ECNU), Shanghai, China.

[8]Department of Neurosurgery, Xuanwu Hospital, Capital Medical University, Beijing, China.

[9]Department of Ophthalmology, National University of Singapore, Singapore.

[10]Singapore Eye Research Institute, Singapore National Eye Centre, Singapore.

[11]Ophthalmology and Visual Science Academic Clinical Program, Duke-NUS Medical School, Singapore.

[12]Department of Pediatrics, The First Hospital of Tsinghua University, Beijing, China.

[13]College of Future Technology, Peking University, Beijing, China.

[14]School of Health and Wellbeing, University of Glasgow, Glasgow, UK.

[15]China National Clinical Research Center for Neurological Diseases, Beijing Tiantan Hospital, Capital Medical University, Beijing, China.

[16]Vanke School of Public Health, Tsinghua University, Beijing, China.

[17]Genetics service, KK Women's and Children's Hospital, Singapore, Singapore.

[18]Bioinformatics Institute (BII), Agency for Science, Technology and Research (A*STAR), Singapore, Singapore.

[19]Rare Care Centre, Perth Children's Hospital, Perth, WA, Australia.

[20]SingHealth Duke-NUS Institute of Precision Medicine, Singapore, Singapore.

[21]Department of Computer Science and Engineering, School of Electronic, Information, and Electrical Engineering, Shanghai Jiao Tong University, Shanghai, China.

[22]MOE Key Laboratory of AI, School of Electronic, Information, and Electrical Engineering, Shanghai Jiao Tong University, Shanghai, China.

*Joint First Authors ** Co-Corresponding Authors

**Abstract**

Rare diseases affect millions of individuals worldwide, yet timely diagnosis remains a major public health challenge due to scarcity of specialized clinical expertise. While large language models (LLMs) show promise to support rare disease diagnosis, current models are constrained by insufficient clinical deployability, limited clinically grounded evidence, and scarcity of training data. Here we present RaDaR (Rare Disease navigatoR), an open-source, compact reasoning LLM (32B parameters) for rare disease diagnosis. RaDaR was trained with 49,170 publicly available free-text cases and 104,666 synthetic cases with reasoning-enhanced training. RaDaR showed the strongest performance among evaluated open-source models, including the 671B DeepSeek-R1, across public benchmarks and four external validation centers. In a retrospective cohort, RaDaR prioritized the final diagnosis before documented clinical suspicion in 61.06% of cases, corresponding to a potential lead time of 1.87 months and 50.18% of the within-centre interval. In a randomized physician-assistance trial, RaDaR assistance improved physicians' rare-disease diagnostic accuracy by 21.44 percentage points compared with internet search alone ($P < 0.0001$). Synthetic-data ablations suggested that phenotype-anchored narratives provide useful training signal for long-tail rare diseases, with a monotonic scaling trend within the tested data range. Together, RaDaR and its development and validation framework provide a deployable rare-disease reasoning model and a reproducible development framework for diagnostic AI under data scarcity.

## Introduction

Rare diseases constitute a critical global health challenge. Though individually rare, collectively, rare diseases affect 400 million patients, constituting nearly 6% of the global population[1-3]. Diagnosis is the central challenge in rare disease care. On average, patients wait 6.2 years to receive a confirmed diagnosis. Nearly half have more than five clinical visits and see six to eight different clinicians on average, a journey often described as a "diagnostic odyssey" [4-10]. The diagnostic odyssey reflects not only the cognitive challenge of recognizing thousands of heterogeneous conditions, but also the reasoning burden in fragmented and incomplete clinical evidence in the context of highly dispersed rare-disease expertise[11,12].

AI-assisted diagnosis has emerged as a promising way to help bridge this knowledge gap by suggesting candidate conditions (usually termed differential diagnoses) beyond the typical physician's regular training and experience. Prevailing methods such as PhenoBrain[13] and Exomiser[14] rely on structured representations like Human Phenotype Ontology (HPO), which is usually constrained by the low-granularity of patient presentations and heavy reliance on physicians for accurate phenotype extraction. The recent emergence of Large Language Models (LLMs) offers a transformative opportunity to overcome these barriers as an alternative approach to diagnostic support[15-17]. LLMs can directly interpret unstructured clinical narratives, leveraging their extensive knowledge and reasoning capabilities to navigate complex presentations of rare diseases. Recent studies have also demonstrated the potential of proprietary LLMs to facilitate rare disease diagnosis in retrospective cases[18-20].

However, translating theoretical efficacy of current LLMs into clinical practice faces substantial barriers. The first is an evaluation gap. Most existing studies assess models' performance on retrospective and static cases, but such evaluations provide limited insight into real-world clinical impact, including when a diagnosis can be suggested, how many patients may benefit from earlier prioritization, or whether physicians would actually change their diagnosis when assisted by the model[18,19,21]. The second is a data gap. Rare diseases are characterized by sparse real-world observations and extreme long-tail distributions, limiting both model training and evaluation. The third is a deployment gap. Current high-performing language models or agentic models are either closed-source (e.g., the GPT-o series), precluding their use under strict patient-data privacy regulations, or possess prohibitive parameter counts and require high computational resources (e.g., 671B parameters for DeepSeek-R1[22]), limiting local deployment in resource-constrained hospital environments[23,24]. These gaps in evaluation, data scarcity, and deployment continue to constrain the responsible clinical adoption of LLM-based diagnostic systems.

To bridge these gaps, we introduced RaDaR (Rare Disease navigatoR), an open-source and 32B-parameter compact LLM tailored for rare disease diagnostics. We evaluated RaDaR across static free-text benchmarks, phenotype-input benchmarks, and multicenter external EHR datasets. To move beyond conventional single-case evaluation, we further tested RaDaR in cumulative multi-visit diagnostic trajectories, estimating whether the model could prioritize the final diagnosis before documented clinical suspicion. We then conducted a randomized physician-assistance trial, comparing conventional internet search with internet search plus RaDaR (Fig. 1). Finally, we evaluated the utility and scalability of the synthetic data construction pipeline. Together, these analyses position RaDaR not only as a deployable rare-disease reasoning model, but also as a claim-proportionate evaluation framework and scalable training paradigm for medical LLMs under data scarcity.

## Results

### An overview of the model development and evaluation

In this study, we developed and released RaDaR, an open-source LLM specialized in rare disease diagnostics, initialized from the DeepSeek-R1-Distill-Qwen-32B backbone. The training corpus integrated real-world case reports from the RareArena dataset[25] (derived from PMC case reports) with synthetic cases generated by powerful LLMs via a phenotype-anchored pipeline. The model underwent a two-stage reasoning enhancement process: (1) Supervised Fine-Tuning (SFT) utilizing 154k cases (49,170 real and 104,666 synthetic cases) with reasoning trajectories distilled from DeepSeek-R1; and (2) Direct Preference Optimization (DPO) using 14k preference pairs on 4,750 selected cases derived from SFT model rollouts (Fig. 1).

We evaluated the model across three distinct benchmarks to assess in-distribution retention, real-world generalization, and adaptability across input format. First, we constructed a RareArena hold-out test set by sampling up to three cases per disease (encompassing all available cases for diseases with 3 or fewer samples), yielding 12,938 cases covering 3,727 diseases. Within RareArena, two tasks were involved. One is rare disease screening task (RDS) evaluating diagnosis intuition by simulating early-stage scenarios where physicians need to identify potential diseases from vague clinical signals. The other is rare disease confirmation (RDC), which assesses diagnosis accuracy with decisive evidence provided[25]. Second, we performed external validation by collecting real-world EHRs of rare disease patients from four centers (detailed demographic characteristics are presented in Extended Data Table. 1). Third, despite our focus on free-text input, we incorporated RareBench (1,196 cases)[26] and Reese et al. (5,213 cases)[27] to assess the model's adaptability to structured phenotype inputs, benchmarking against established phenotype-based diagnostic systems.

We benchmarked our model against 1) leading open-source general LLMs, including DeepSeek-R1[22] and gpt-oss[28]; 2) latest LLMs featuring advanced clinical reasoning capacities, including HuatuoGPT-o1[29], m1[30], and Baichuan-M2[31]; 3) phenotype-centered diagnostic systems, including PhenoBrain[13], Exomiser[14], PubCaseFinder[32] and LIRICAL[33].

**Cross-sectional diagnostic evaluation of RaDaR on public and external datasets**

Our model, RaDaR, demonstrated the strongest overall diagnostic performance among evaluated open-source models across internal, external, and public benchmarks, indicating its specialized clinical reasoning capabilities and supporting its potential as a clinical decision-support tool (Fig. 2). We used top-5 recall as the primary diagnostic performance metric, reflecting the intended clinical use of RaDaR: to assist clinicians by prioritizing a short differential diagnosis list rather than replacing clinical judgment with a single definitive answer. The output length was fixed at five to balance coverage and clinical usability: it is sufficiently broad to capture plausible alternatives while remaining concise enough for clinicians to review. Top-1 and top-3 recall were additionally reported to provide a more comprehensive characterization of model performance (Extended Data Table. 2).

First, in the RareArena hold-out evaluation, the model ranked first for open-source diagnostic systems, achieving 42.53% diagnostic accuracy in RDS. The performance surpassed both the leading open-source general LLMs (gpt-oss-120b: 38.40%; DeepSeek-R1: 35.52%) and the top-tier medical-domain model, Baichuan-M2 (28.53%). In RDC, it secured the highest diagnostic accuracy (70.34%) among all evaluated models. Notably, this performance advantage was achieved despite RaDaR's relatively compact size, supporting its potential for local clinical deployment (Fig. 2a). We explored RaDaR's reasoning traces and found that they generally followed a clinically recognizable diagnostic workflow: summarizing key clinical findings, generating differential diagnoses, and analyzing the likelihood of each possibility (examples shown in Appendix A-C).

We also evaluated the model's performance across different disease subgroups. In terms of prevalence, RaDaR consistently outperformed all other models in RDS tasks across disease subgroups with varying prevalence rates (Extended Data Fig. 1). Its advantage was most pronounced in the rarest cases, where it surpassed the second-best model by 1.07% for diseases affecting 1–5 per 10,000 people, and by 6.90% for ultra-rare diseases affecting 1 per 1,000,000. In RDC, though not achieving the best performance in every prevalence subgroup, RaDaR had performance comparable to that of DeepSeek-R1 and gpt-oss. We next assessed performance across different clinical specialties (Extended Data Fig. 2,3). In RDS, RaDaR consistently outperformed all open-source comparators across every specialty. In RDC, this advantage was preserved for most specialties, with a modestly lower performance in rare hepatic and respiratory diseases. Finally, stratification by genetic status (genetic vs non-genetic) and multisystem involvement (present vs absent) showed a consistent pattern, with RaDaR outperforming baseline models across all subgroups in both tasks (Extended Data Fig. 4).

Second, external validation on multicenter real-world patient records supported RaDaR's robustness under distributional shifts. Despite substantial heterogeneity in real-world clinical documentation, RaDaR achieved a diagnostic accuracy of 66.92% (± 0.35) on the PUMCH (Peking Union Medical College Hospital) dataset, outperforming DeepSeek-R1 (61.80% ± 0.46) and gpt-oss-120b (62.13% ± 0.63). Similarly, RaDaR attained the highest accuracy on XHCMU (Xuanwu Hospital, Capital Medical University), XJH (Xijing Hospital) and KKH (KK Women's and Children's Hospital) datasets with 72.57%, 84.07% and 48.08% accuracy respectively. Collectively, these results support RaDaR's generalizability beyond the public case-report training distribution and underscore its potential for local clinical deployment (Fig. 2b, Extended Data Table. 2).

Third, in phenotype-input benchmark evaluations, RaDaR demonstrated strong versatility despite being optimized for free-text clinical records (Fig. 2c, Extended Data Table. 2). On RareBench, RaDaR achieved the second-best diagnostic accuracy of 53.93%, outperforming all LLMs except gpt-oss (55.02%) and all traditional phenotype-centered tools. On the Reese et al. benchmark, RaDaR achieved the highest accuracy (48.17%), outperforming both LLM-based models and phenotype-centered tools.

**Free-text reasoning complements phenotype-centered rare-disease diagnosis**

Because rare-disease diagnosis has traditionally relied on phenotype extraction and ontology-based prioritization, we directly compared three diagnostic strategies: traditional phenotype-centered tools, phenotype-centered agent systems, and free-text reasoning approaches (Extended Data Table. 3). Across the three external validation datasets, full-text RaDaR achieved the best overall performance, exceeding the phenotype-centered agent system DeepRare by 34.69, 35.43, and 21.11 percentage points in PUMCH, XHCMU, and XJH respectively. Case-level analysis suggested that this performance gap was partly attributable to inaccurate phenotype extraction and the loss of key diagnostic information during phenotype encoding (Appendix D). Consistently, RaDaR performed substantially worse when restricted to phenotype-only inputs than when provided with full-text clinical records in external validation. Phenotype-only input reduced RaDaR's performance from 66.92% to 35.07% in PUMCH, from 72.57% to 28.57% in XHCMU, and from 84.07% to 55.56% in XJH. These findings should not be interpreted as diminishing the value of phenotype-centered methods; rather, they suggest that free-text reasoning can complement ontology-based diagnostic frameworks by preserving narrative clinical context, particularly in real-world clinical settings where phenotypes are incompletely documented or variably encoded.

**RaDaR prioritizes final diagnoses earlier along retrospective diagnostic trajectories**

Despite RaDaR's satisfactory performance on cross-sectional case-level evaluation, its potential value along diagnostic workflows requires clinically grounded evaluation. We therefore conducted two studies to systematically characterize its clinical potential. First, to estimate whether RaDaR could prioritize the final diagnosis earlier along real diagnostic trajectories, we performed a retrospective cumulative-visit evaluation of patients with confirmed rare diseases and multiple prior undiagnosed encounters. We applied RaDaR at each clinical encounter to generate a differential diagnosis list with no more than 10 diseases. The visit at which clinicians first documented suspicion of the final rare disease served as the physician baseline. Reviewers then manually determined the earliest visit at which RaDaR correctly identified the final diagnosis, enabling a direct comparison of whether, and by how many visits, the model prioritized the final diagnosis before documented clinical suspicion (Fig. 3a). Importantly, this evaluation was not designed to test longitudinal reasoning per se, but rather to provide a clinically interpretable estimate of the potential diagnostic benefit of RaDaR in real-world diagnostic trajectories.

A total of 113 patients were enrolled (86 from PUMCH and 27 from XHCMU). The cohort included 50 male and 63 female participants, with the mean age of symptom onset at 31.07 ± 22.00 years (Extended Data Table. 4). Genetic diseases accounted for 32.74% of the cohort, and 49.56% of cases involved multiple systems. Following previous work[34], the diagnostic trajectory was divided into two distinct intervals: (1) the pre-center interval, defined as the time from symptom onset to the initial visit at a participating center; and (2) the within-center interval, defined as the time from that first visit to the point of clinical suspicion. The total diagnostic duration was calculated as the sum of these two intervals, encompassing the entire period from symptom onset to diagnostic suspicion. The mean total diagnostic duration was 46.14 ± 70.36 months. Specifically, the pre-center interval lasted 40.36 ± 68.91 months, and 95.58% of patients had sought medical consultations at other institutions before presenting to the participating centers. Even within these centers, patients underwent an average of 4.46 ± 4.17 clinical visits over 5.78 ± 12.37 months before diagnostic suspicion was established.

RaDaR outperformed the gpt-oss and DeepSeek-R1 in this evaluation. Using the visit-normalized diagnostic advancement score (VNDAS), which quantifies how much earlier a model can identify the correct diagnosis along the diagnostic trajectory, RaDaR achieved a score of 0.47, compared with 0.41 for gpt-oss and 0.37 for DeepSeek-R1 (Fig. 3b). RaDaR correctly identified the final diagnosis before clinical suspicion in 61.06% of cases and at the same visit in 25.66%, while failing to identify the diagnosis in only 13.27% of cases, outperforming both gpt-oss and DeepSeek-R1 (Fig. 3c). Within this retrospective framework, RaDaR prioritized the final diagnosis 1.62 visits before documented clinical suspicion (95% CI: 1.22–2.02), corresponding to a potential lead time of 1.87 months (95% CI: 0.55–3.21). When normalized at the patient level by each individual's within-centre diagnostic interval, this translated into a mean patient-level lead-time ratio of 50.18% (95% CI: 41.42–58.93). As illustrated by the heatmaps (Fig. 3d), although most early detections clustered within the immediate 1–2 visits preceding the clinical diagnosis, a subset of patients showed a markedly longer lead time. Specifically, 11.50% (13/113) of the cohort had a correct model prediction more than 3 months before clinical suspicion, and 4.42% (5/113) more than 10 months in advance. In terms of clinical encounters, RaDaR prioritized the diagnosis more than three visits earlier in 23.89% (27/113) of patients and by more than six visits in 3.54% (4/113) of patients. Considering that physicians may not be able to evaluate all differential diagnoses suggested by the model, we also reported performance under constrained settings in which only the top one, three, and five differential diagnoses were considered (Extended Data Table. 5).

We further characterized the flow of clinical information throughout the diagnostic journey and explored which types of information changes were associated with diagnostic transitions. As illustrated by a representative patient, clinical information was dynamically accumulated and updated over time: symptoms and laboratory findings were more frequently newly added, whereas treatments and laboratory findings were more often updated (Extended Data Fig. 5). We next investigated which types of clinical information changes were associated with incorrect-to-correct diagnostic transitions. Each additional newly added imaging or procedural item was significantly associated with an incorrect-to-correct diagnostic transition (OR = 1.15, $P = 0.030$), whereas no significant associations were observed for other information types ($P > 0.05$). We further conducted subgroup analyses to assess the stability of model performance across disease categories. Model performance, measured by both diagnostic advancement score and diagnostic advancement ratio, was not significantly associated with disease etiology (both $P > 0.05$; Extended Data Fig. 6).

**RaDaR assistance improves physicians' rare-disease diagnostic accuracy**

To evaluate whether RaDaR can assist physicians in an interactive manner, we conducted a randomized physician-assistance trial comparing RaDaR-assisted diagnosis with a conventional internet-search strategy. Participating physicians from neurology, internal medicine, and pediatrics—three specialties in which rare diseases are most frequently encountered—were asked to diagnose five department-specific cases. Participants were randomized at the physician level to one of two arms: a control group with access to the internet alone, and a RaDaR-assisted group with access to both RaDaR and the internet. In total, 84 physicians from 28 hospitals were enrolled and randomized, of whom 76 completed the trial. Among those who completed the trial, 40 were assigned to the control arm and 36 to the RaDaR-assisted arm (Extended Data Fig. 7). The median clinical experience was 4 years (interquartile range [IQR], 2–9). Baseline characteristics are summarized in Extended Data Table. 6.

Baseline unassisted diagnostic performance did not differ significantly between the control and RaDaR-assisted arms (11.50 ± 14.24 vs. 13.33 ± 16.56, $P = 0.60$). In the assisted setting, however, physicians in the RaDaR-assisted group achieved significantly higher diagnostic accuracy than those in the control group (49.44 ± 22.67 vs. 28.00 ± 22.10; absolute difference, 21.44; 95% CI, 11.20–31.68; $P < 0.0001$). Diagnostic time did not differ between groups (control: 3.69 ± 3.12 minutes; RaDaR: 3.82 ± 2.36 minutes; absolute difference, 0.14; 95% CI, -1.14–1.41; $P = 0.8332$). Self-reported helpfulness was higher with RaDaR than with conventional search (4.19 ± 0.63 vs. 3.36 ± 0.80; absolute difference, 0.83; 95% CI, 0.50–1.16; $P < 0.0001$) (Fig. 4).

We next examined whether the benefit of RaDaR varied across physician subgroups. Descriptively, RaDaR-assisted diagnosis showed higher diagnostic accuracy than the control arm across physician-experience strata and in most specialty subgroups (Fig. 4e, f). No significant interaction was observed between intervention group and physician experience (OR = 0.84, $P = 0.695$) or department, including internal medicine (OR = 0.90, $P = 0.827$) and neurology (OR = 0.35, $P = 0.135$), with pediatrics as the reference group.

We also conducted an exploratory safety-related analysis of physician confidence and diagnostic calibration. Self-reported confidence was collected after each case and calibration error was defined as the discrepancy between confidence and accuracy (details in Methods). We found that calibration error was significantly higher in the RaDaR-assisted arm during the assisted phase (0.48 ± 0.15 vs. 0.39 ± 0.17, $P = 0.02$). A case-level descriptive analysis further showed that, among incorrectly diagnosed cases, high or very high confidence scores (4 or 5) were more frequent in the RaDaR-assisted arm than in the control arm (65.9% vs. 27.1%). Among correctly diagnosed cases, both arms showed similarly high confidence levels. These findings indicate that improved diagnostic accuracy may be accompanied by a calibration trade-off, with incorrect diagnoses more frequently accompanied by high confidence (Extended Data Fig. 8).

**Synthetic narratives mitigate data scarcity and support performance scaling**

To systematically quantify the contribution of our synthetic data and to explore the potential of the data synthesis pipeline, we conducted a series of ablation studies using a scaled-down model (DeepSeek-R1-Distill-Qwen-14B). Specifically, we benchmarked three distinct training paradigms: the "real-world" setting (trained exclusively on real-world case reports), the "real-world + synthetic" setting (trained on real-world case reports supplemented by synthetic data), and the experimental "synthetic-only" setting (trained solely on synthetic data).

First, we evaluated the impact of data augmentation. Compared to the "real-world" setting, the model augmented with synthetic data showed a consistent performance increase (Fig. 5a). The performance gain showed an inverse pattern with the availability of authentic data, and the effect of our approach was most pronounced in the ultra-rare regime (Bin "0"). In this data-sparse scenario, the inclusion of synthetic data yielded a substantial improvement of +7.31 percentage points in diagnostic accuracy (Fig. 5b). As visualized in Extended Data Fig. 9, the injection of synthetic data effectively transforms the original extreme long-tail distribution into a more balanced distribution. In addition to the availability of authentic data, we also examined the training-data distribution and model performance across disease-prevalence groups (Extended Data Fig. 9 and Fig. 10). Authentic data covered only a small fraction of ultra-rare diseases (3.5% in diseases with prevalence <1/1 000 000), while incorporating synthetic data substantially expanded the disease coverage to 32.84%. Correspondingly, we observed the largest performance gain in the ultra-rare subgroup, with an improvement of 7.32% in diagnostic accuracy. This suggested that incorporating synthetic data may help alleviate performance disparities caused by limited coverage and skewed training distributions.

Notably, under the "synthetic-only" paradigm, we observed a notable result: the model trained solely on synthetic data achieved a diagnostic accuracy of 39.08%, comparable to "real-world" setting (37.58%) trained on authentic case reports, suggesting that phenotype-anchored synthetic narratives can provide useful training signal (Fig. 5c). Similar performance patterns were also observed across three external datasets, suggesting that the utility of synthetic-data training may extend to external clinical datasets (Extended Data Fig. 11). Furthermore, even for the subset of diseases where synthesis was infeasible due to missing Orphanet annotations, the model demonstrated non-trivial diagnostic generalization. While performance on these non-synthesized conditions was naturally lower, likely reflecting their inherent obscurity, the model still exhibited significant improvement compared to the baseline, suggesting potential transfer to diseases not directly represented in the synthetic-data generation process (Fig. 5d).

Finally, we examined the scalability of this approach by varying the amount of synthetic training data. Across the tested range of 1 to 30 synthetic samples per disease, diagnostic performance increased monotonically as synthetic-data volume grew and followed a positive power-law scaling trend (scaling exponent, 0.082; 95% CI, 0.070–0.095; $R^2$ = 0.99 for RareArena; scaling exponent, 0.056; 95% CI, 0.013–0.100; $R^2$ = 0.85 for external validation datasets) (Fig. 5e, Extended Data Fig. 11). Although computational constraints precluded further extrapolation, this trend suggests a scalable path for improving model performance through further expansion of synthetic training data.

**Discussion**

In this study, we developed and released RaDaR, a specialized open-source compact LLM for rare disease diagnosis. The model outperformed the evaluated open-source models, including DeepSeek-R1, in both public benchmarks and four external validation centers. More importantly, RaDaR provided evidence of potential diagnostic lead time in a retrospective cohort study and demonstrated potential to support physicians' clinical diagnosis in a randomized physician-assistance trial. Finally, the synthetic data construction pipeline has demonstrated its utility, transferability, and scalability, providing a framework for future rare disease LLM development under data scarcity.

Many rare disease diagnostic pipelines, ranging from Exomiser-like tools to recent agentic frameworks, rely heavily on phenotype extraction and ontology-based term standardization for disease prioritization. As a result, their performance is inherently sensitive to both the accuracy of phenotype extraction and the completeness of existing disease–phenotype knowledge bases. In contrast, RaDaR directly processes free-text clinical records and demonstrates generalizability to diseases with no phenotype mapping information (Fig. 5d), allowing it to better adapt to the complex real-world clinical scenarios. The results do not imply that phenotype-centered approaches are inherently inferior. Indeed, LIRICAL achieved the highest top-1 recall on RareBench among all evaluated methods, including all LLMs, demonstrating the potential of phenotype-centered approach, consistent with previous studies[27]. Rather than replacing phenotype-based methods, our findings suggest that integrating phenotype-centered approaches with free-text reasoning may represent a promising direction for future rare-disease diagnostic systems. Some studies have already explored the potential of LLMs to improve automated phenotyping[35,36] Similarly, we believe RaDaR could complement the current rare disease diagnostic paradigm through its free-text processing and reasoning capacity, and may serve as a flexible reasoning module that can be integrated into ontology-based or agent-based rare disease diagnostic workflows.

Beyond RaDaR’s diagnostic performance, our work addresses the key challenges limiting the application of large language models in rare diseases: insufficient clinical deployability, limited evidence of clinical utility, and scarcity of training data. First, RaDaR's 32B-parameter scale enables efficient in-hospital deployment while delivering performance that rivals or even exceeds much larger general-purpose models. Unlike prohibitive ultra-large models that are often inaccessible to resource-constrained clinical settings, RaDaR offers a stronger performance–efficiency trade-off, broadening access to high-quality diagnostic decision support. Moreover, the open-source design of RaDaR also supports privacy-preserving local deployment, allowing hospitals to maintain greater control over data governance and model use[37-39].

Second, RaDaR provided more clinically grounded evidence of its potential utility than previous work. Prior studies have predominantly evaluated models using static, single-encounter records from a single center, an approach that facilitates benchmarking but offers limited insight into real-world clinical utility. In our study, we evaluated RaDaR at multiple time points along a dynamic and iterative diagnostic trajectory, offering a clinically interpretable estimate of potential diagnostic lead time (Fig. 3). We found that RaDaR indeed demonstrated substantial promise for earlier diagnostic prioritization. For example, in a patient with a 16-year history of left-sided lipomatous overgrowth and negative results from prior peripheral-blood whole-exome sequencing and mitochondrial testing, RaDaR proposed “PIK3CA-related overgrowth spectrum” at the first encounter. This hypothesis was subsequently supported by genetic testing performed on a soft-tissue biopsy several months later, after physicians considered the diagnosis at the fourth clinical visit. The diagnostic challenge in this case lies in the somatic nature of the disease-causing mutation, which is frequently missed by routine peripheral-blood exome sequencing. If such prioritization were acted upon in clinical practice, clinicians might consider targeted testing earlier, thereby potentially shortening the diagnostic journey. Further, while the absolute cohort size is constrained, it is 15-fold larger than that of the previous studies employing comparable validation strategies[40]. To move beyond retrospective cohort validation, which evaluates model performance but cannot assess whether clinicians would use model suggestions in iterative real-world decision-making, we conducted a randomized physician-assistance trial showing that RaDaR improved physicians' rare-disease diagnostic performance. We also revealed potential safety signals, most notably an increase in calibration error, that warrant caution in future deployment. Because rare-disease diagnosis is protracted and eligible patients are intrinsically scarce [41,42], an RCT fully embedded in routine clinical care could take years to mature. Our current clinical validation provides an early and practical view of both the promise and the guardrails required for RaDaR-enabled rare-disease diagnosis.

Third, beyond clinical deployment, we developed a data synthesis pipeline that showed utility and transferability to future LLM training. By translating structured knowledge bases into clinically realistic reasoning narratives, we generated phenotype-anchored synthetic cases and found that a model trained exclusively on synthetic data achieved performance comparable to that of a model trained solely on real-world case reports. This finding suggests that phenotype-anchored synthesis is not merely a data augmentation strategy, but a practical solution to mitigating the inherent scarcity and extreme long-tail distribution of rare disease data. Furthermore, the observed scaling trends suggest that increasing the volume of high-quality synthetic data provides a predictable path to enhancing performance. This paradigm effectively decouples dependence on large real-world datasets, offering a scalable and privacy-preserving roadmap for supporting more accessible model development across diverse, data-scarce clinical domains.

**Limitations**

Our study has several limitations. First, despite combining synthetic and real-world case reports and covering 5,877 diseases, nearly 2,000 rare diseases remain unrepresented in the training set due to the absence of phenotype–disease knowledge for ultra-rare conditions. Future work should expand knowledge coverage and develop alternative strategies to fill these gaps. Second, the current model is purely language-based and does not incorporate multimodal clinical data, which may provide more detailed information for rare disease diagnosis. Third, owing to the highly fragmented nature of the rare disease diagnostic journey and limited EHR interoperability, reconstructing a patient's complete trajectory across multiple institutions remains a significant challenge. In this study, data derived from top-tier tertiary centers represent only the final stage of a much longer diagnostic process. Consequently, the true 'add-on' effect of RaDaR may be underestimated if deployed earlier and more broadly across the healthcare ecosystem. The constraints we encountered underscore the urgent need for integrated rare-disease data infrastructures in future research. Fourth, since several benchmark cases were derived from public case reports, and the base model was pretrained on large-scale public corpora, benchmark contamination cannot be fully excluded. We attempted to mitigate this risk by using rephrased benchmark narratives (RareArena) and evaluating non-public external EHR cohorts. Nevertheless, these analyses cannot guarantee complete absence of pretraining exposure. Future validation should prioritize prospectively collected, non-public and multi-institutional cases. Fifth, our benchmark evaluation relied on three-level or binary scoring paradigms, which may not fully capture the nuances of model performance. Future work should develop more comprehensive, precise, and automated evaluation frameworks for free-text diagnostic outputs. Sixth, although we incorporated a multi-visit EHR evaluation, we did not explicitly incorporate longitudinal modeling into the training process. This remains an important direction for future exploration. Seventh, while our retrospective analyses and RCT supported RaDaR's clinical potential, whether this translates into meaningful clinical benefit, including changes in testing, referral, treatment, or patient outcomes, requires rigorous, long-term prospective validation. The randomized physician-assistance trial was a controlled case-based study rather than a pragmatic trial embedded in routine clinical workflows, and the exploratory calibration findings highlight the need for safeguards against overconfidence and automation bias before deployment.

## Conclusion

In conclusion, we developed and released RaDaR, a specialized open-source and compact 32B LLM for rare disease diagnosis, achieving leading performance across public benchmarks and multi-center real-world cohorts. Our clinical evaluation demonstrated RaDaR's potential to prioritize final diagnoses earlier along retrospective diagnostic trajectories and to support physicians' diagnostic performance, as supported by a randomized physician-assistance trial. In parallel, our phenotype-anchored synthetic data pipeline offers a scalable and reproducible solution to addressing rare disease data scarcity and long-tail imbalance. Beyond developing a deployable reasoning model with clinically grounded evidence of potential, this work offers a reproducible and scalable paradigm for building future clinical reasoning models in data-scarce settings.

**Data availability**

Simulated phenotype sets were generated using publicly available data from Orphanet (https://www.orpha.net/) and the Human Phenotype Ontology (HPO; https://hpo.jax.org/). The RareArena dataset used for both model training and benchmarking is open-source and publicly available (https://github.com/zhao-zy15/RareArena). The two phenotype-based rare disease benchmarks are also publicly available: RareBench (https://github.com/chenxz1111/RareBench) and the dataset described by Reese et al. (https://zenodo.org/records/15324355). Due to privacy concerns, the in-house datasets used for external validation and retrospective analyses are not publicly available. Access may be considered upon reasonable request to the corresponding author with a clear research proposal and data-use plan. The processed analysis data and de-identified raw data from the randomized physician-assistance trial are also available from the corresponding author upon reasonable request.

**Code availability**

The model weights and full training datasets are both available on Hugging Face (https://huggingface.co/sczzz/RaDaR-32B). The training scripts and synthetic-data construction code are available on GitHub (https://github.com/sczzz3/RaDaR). The web application developed for this study is hosted at https://raredx.datummed.com.

**Contributors**

TYW had full access to all of the data in the study and took responsibility for the integrity of the data and the accuracy of the data analysis. HC, SZ, and ZZ conceived and planned this study. HC, SZ, and ZZ drafted the manuscript. HC and SZ did the data synthesis. SZ trained the model. HC, SZ, and ZZ did data interpretation. HC, XJ, JZ and SH organized the external validation and retrospective cohort study. SH, XJ and SM evaluated the case quality. HC, XJ, JZ, SH, SM and DZ designed and participated the randomized AI physician-assistance trial. All authors contributed to the intellectual development of this study and provided critical feedback on the manuscript. All authors had full access to all the data in the study and had final responsibility for the decision to submit for publication.

**Declaration of interests**

TYW declares consulting fees from Aldropika Therapeutics, Bayer, Boehringer Ingelheim, Genentech, Iveric Bio, Novartis, Plano, Oxurion, Roche, Sanofi, and Shanghai Henlius; funding from the National Key R&D Program, China (grant number 2022YFC2502802); and being an inventor, patent holder, and cofounder of the start-up companies EyRiS and Visre. All other authors declare no competing interests.

**Acknowledgments**

TYW received funding from National Key R & D Program of China (2022YFC2502802), National Natural Science Fund of China (8238810007) and Beijing Natural Science Foundation (IS23096). SY received funding from the Natural Science Foundation of China (12171270). SYZ received funding from National Key R &D Program of China (2022ZD0116001). The GPU computing resources were supported by the High-Performance Computing Center of Tsinghua University.

## Method

### Datasets

In this study, we used several publicly available rare disease benchmarks (both free-text based and phenotype-based) and also in-house rare disease cases to evaluate the performance of the baseline methods and RaDaR.

*RareArena*[25] is a comprehensive rare disease diagnostic dataset based on free-text modality comprising 49,760 cases for the rare disease screening (RDS) task and 22,901 cases for the rare disease confirmation (RDC) task, covering more than 4,000 diseases. There are two subtasks: Rare

Disease Screening (RDS), where the cases are truncated up to any diagnostic tests, such as whole-genome sequencing for genetic diseases and pathogen detection for rare infections, and Rare Disease Confirmation (RDC), where the cases are truncated up to the final diagnosis.

*RareBench*[26] is a phenotype-based benchmark designed to systematically evaluate the capabilities of LLMs from four distinct perspectives in the context of rare diseases. In this work, we focused specifically on Task 4: Differential Diagnosis among Universal Rare Diseases.

*Reese et al*[27]. is another phenotype-based benchmark comprised of 5,213 solved rare disease cases, along with basic demographic and onset information. The cohort covers 378 Mendelian and chromosomal disorders curated from 726 published studies. Only phenotype data were used for analysis, providing a standardized benchmark for evaluating rare-disease diagnostic methods.

*In-house dataset*. Electronic health records (EHRs) from four tertiary centers were collected: Peking Union Medical College Hospital (PUMCH), Xuanwu Hospital, Capital Medical University (XHCMU), Xijing Hospital (XJH) and KK Women's and Children's Hospital (KKH). After removing duplicates, we retrospectively collected inpatient admission records from PUMCH (September 2024 to July 2025), XHCMU (February 2025 to May 2025), XJH (February 2025 to June 2025) and KKH (January 2015 to December 2020). Patients were included if they had a confirmed rare disease diagnosis. Diagnoses were manually extracted and annotated by clinical experts. To prevent label leakage, if the diagnosis explicitly appeared in the EMR, we truncated the record immediately before its first explicit mention and used only the preceding text for evaluation. If no explicit diagnosis appeared in the admission record, we used the full record.

**Model performance evaluation**

For RareArena, we largely followed the original evaluation protocol. In the original RareArena benchmark, predictions were scored on a three-level scale: 2 for an exact or synonymous match, 1 for a hypernym match, and 0 for an unrelated answer. Scoring was performed automatically by GPT-4o, which demonstrated satisfactory agreement with two clinician reviewers, with kappa values of 0.95 and 0.97 for top-1 scoring and 0.83 and 0.87 for top-5 scoring. These values were comparable to the inter-clinician agreement (0.89)[25].

Here, we made the hypernym criterion more stringent. Rather than accepting broad parent disease categories, we restricted valid hypernyms using the Orphanet classification file

(https://sciences.orphadata.com/classifications/). A prediction was counted as a hypernym only when it corresponded to an upper-level disease concept with fewer than 10 daughter disorder concepts. For example, for “Long chain 3-hydroxyacyl-CoA dehydrogenase deficiency (ORPHA:5)”, “3-hydroxyacyl-CoA dehydrogenase deficiency (ORPHA:309127)” can be accepted as valid hypernyms, whereas “Metabolic disease with cataract (ORPHA: 98644)” cannot. Answers that were more specific than the reference diagnosis were considered unrelated and assigned a score of 0. The updated scoring prompt is provided in Appendix J, and the Orphanet-based hypernym mapping file is available on GitHub (https://github.com/zhao-zy15/RareArena).

For the in-house validation datasets, real-world clinical diagnoses were not always directly mappable to Orphanet codes, making hypernym-based scoring hard. We therefore adopted a binary evaluation paradigm. A prediction was considered correct only if it exactly matched, or was synonymous with, the real-world reference diagnosis; otherwise, it was considered incorrect. The corresponding evaluation prompt is provided in Appendix K.

## Baselines

### Phenotype-based diagnostic tools

*PhenoBrain*[13] is a phenotype-based diagnostic pipeline that extracts structured clinical phenotypes from electronic health records using a BERT-based model and ranks candidate rare diseases via an ensemble of specialized predictors, serving as a strong non-LLM baseline for differential diagnosis of rare diseases.

*Pubcasefinder*[32] is a phenotype-driven rare disease diagnostic system that matches patient phenotypes encoded as HPO terms against curated case reports and disease databases, ranking candidate rare diseases based on phenotype similarity.

Exomiser[14] prioritizes candidate disease-causing variants from whole-exome or whole-genome sequencing data by combining phenotype data with variant filtering and network analysis. It annotates and filters variants based on frequency and predicted pathogenicity, then ranks genes by integrating phenotype similarity via HPO comparisons and model organism data and network proximity scores. We used the phenotype-only mode for benchmarking.

LIRICAL[33] prioritizes candidate diseases through an interpretable likelihood-ratio framework using HPO-coded phenotypic abnormalities. It estimates how each clinical feature contributes to the support for or against each diagnosis and reports ranked candidate diseases with likelihood-based scores and posterior probabilities.

**Advanced large language models**

*DeepSeek-R1*[22] is an open-source reasoning-focused large language model trained with large-scale reinforcement learning with 671 billion parameters. It exhibits strong performance on tasks requiring mathematical reasoning and structured problem solving.

*gpt-oss*[28] is the OpenAI's open-weight large language models trained with instruction tuning for general-purpose language understanding and generation. It serves as a strong open-source baseline for evaluating reasoning and generation performance across a wide range of tasks. We used the 120 billion parameter version for benchmarking and set the reasoning effort as "high".

*HuatuoGPT-o1*[29] is a series of open-source medical LLMs fine-tuned for advanced medical reasoning. It is designed to generate extended and reflective reasoning before answering for complex medical problems. We used the 72 billion parameter version, which is fine-tuned on Qwen2.5-72B.

m1[30] is a 32B-parameter open-source medical LLM fine-tuned with only 1,000 high-quality medical reasoning examples and enhanced via a test-time scaling strategy that extends inference-time thinking.

Baichuan-M2[31] is a 32-billion-parameter open-source language model fine-tuned for medical-enhanced reasoning. Built on top of Qwen2.5-32B with a novel Large Verifier System and multi-stage reinforcement learning, it targets real-world clinical decision-making and domain-specific inference while retaining strong general capabilities.

**Evaluation of diagnostic approaches across phenotype- and text-based paradigms**

To evaluate the performance of different technical approaches in AI-facilitated rare disease diagnosis, we compared phenotype-based and text-based diagnostic paradigms. The phenotype-based paradigm included both traditional phenotype-centered tools and phenotype-centered agent systems.

For the traditional phenotype-centered approach, clinical phenotypes were first extracted from case descriptions using PhenoTagger and then used as input for phenotype-based diagnostic tools, including PhenoBrain, PubCaseFinder, Exomiser and LIRICAL, to retrieve candidate diagnoses.

For the phenotype-centered agent system, we included DeepRare as a representative baseline for rare disease diagnosis[20]. DeepRare uses a multi-agent LLM framework to integrate clinical phenotypes, HPO terms and, when available, genomic information, producing a ranked list of candidate diseases with evidence-supported reasoning. In our experiments, we followed the original DeepRare configuration, including all default settings and the use of DeepSeek-V3 as the LLM backbone. However, two components could not be fully replicated. First, PubCaseFinder was temporarily unavailable during our evaluation and was therefore excluded (May, 2026). Second, for similar-case retrieval, the original authors only released the RareArena RDS embedding dataset; therefore, the case bank was restricted to this dataset. DeepRare was not evaluated on RareArena because its per-case reasoning time was approximately 5–10 minutes, making large-scale evaluation across more than 10,000 cases computationally impractical.

For the text-based paradigm, we included RaDaR as the representative model, given its strong performance across multiple rare disease diagnostic benchmarks.

## Real-world training data selection

RareArena encompasses a large number of valuable real-world rare disease cases for model development. Thus, we meticulously selected a subset of RareArena as the training set. We first randomly selected up to three cases per disorder and subgroup disorder as validation tests and used the remaining cases for training. We denote the curated real-world training set as $\mathcal{D}_{\mathrm{real}}$.

## Synthetic data curation

However, we observed that the number of real-world cases per rare disease follows a highly long-tailed distribution, which can hinder the model's ability to learn reliable diagnostic patterns for underrepresented diseases. To alleviate this imbalance, we leveraged authoritative rare-disease knowledge bases—Orphanet and the Human Phenotype Ontology (HPO) —together with an advanced LLM to synthesize realistic rare-disease case summaries. These synthetic cases were designed to improve diagnostic coverage for long-tail rare diseases.

Orphanet provides curated disease--phenotype associations annotated with qualitative frequency labels. For a target disease $y$, we constructed a synthetic phenotype set through the following streamlined multi-stage procedure. Let $\mathcal{P}(y)$ denote the set of all non-excluded Orphanet phenotypes associated with disease $y$.

**Primary clinical anchors** ($\mathcal{P}_{\mathrm{prim}}$)

From phenotypes labeled as *Obligate*, *Very frequent*, or *Frequent*, we uniformly sampled (without replacement) up to three phenotypes to form $\mathcal{P}_{\mathrm{prim}}$. A small-size cap was enforced when the phenotype candidate pool is overly small.

**Frequency-aware sampling** ($\mathcal{P}$)

For each remaining phenotype in $\mathcal{P}(y) \setminus \mathcal{P}_{\mathrm{prim}}$, we converted its qualitative frequency label into an inclusion probability and performed Bernoulli sampling, yielding the set $\mathcal{P}$.

**Hypernym abstraction** ($\mathcal{P}_{\mathrm{hyper}}$)

To model imprecise clinical descriptions, we replaced a small number (at most three, scaled to $|\mathcal{P}|$) of phenotypes in $\mathcal{P}$ with their immediate HPO parents, forming $\mathcal{P}_{\mathrm{hyper}}$.

**Differential-diagnosis noise** ($\mathcal{P}_{\mathrm{noisy}}$)

We injected a few confusing phenotypes sampled from diseases that share phenotypes with $y$, replacing their originating phenotypes to obtain $\mathcal{P}_{\mathrm{noisy}}$.

**Set-level dropout** ($\mathcal{P}_{\mathrm{keep}}$)

Let $\mathcal{P}_{\mathrm{left}}$ denote the phenotype set after hypernym replacement and noise injection. To emulate incomplete clinical documentation, we applied a size-dependent set-level dropout by uniformly subsampling $\mathcal{P}_{\mathrm{left}}$ with a retention ratio $r(n)$, where

$$r(n) \sim \begin{cases} \{1.0\}, & n < 10, \\ \{1.0,\ 0.8\}, & 10 \leq n < 20, \\ \{1.0,\ 0.8,\ 0.5\}, & n \geq 20, \end{cases}$$

with each value chosen uniformly from the corresponding set. The resulting phenotype set was denoted as $\mathcal{P}_{\text{keep}}$.

The final phenotype set for a synthetic case was constructed as

$$\mathcal{P}_{\text{final}} = \mathcal{P}_{\text{prim}} \cup \mathcal{P}_{\text{hyper}} \cup \mathcal{P}_{\text{noisy}} \cup \mathcal{P}_{\text{keep}}.$$

Age-of-onset information consistent with Orphanet annotations was incorporated by mapping coarse onset categories to specific age values via rule-based randomization to enhance narrative diversity. In our experiments, we synthesized 30 phenotype sets per disease using the above protocol and conducted ablations by varying the number of generated cases per disease.

With the sampled and preprocessed phenotype set, we prompted GPT-4o to compose full case narratives. Specifically, we employed a two-round generation procedure. At the first round, we asked the LLM to generate only conditioned on the sampled phenotype sets, with a special focus on the primary clinical anchors $\mathcal{P}_{prim}$ to simulate the scenario that physicians would normally note key presentations of a certain patient. We did not tell the LLM the true rare disease name to reduce the risk that it would introduce label-specific features and make the case summary overly prototypical. Then, in the second round, we provided the LLM with the disease label for it to reflect and correct any unreasonable descriptions that have been generated to ensure the veracity of the rare disease case. We design a two-round generation process to explicitly balance diversity and clinical fidelity. During the first round, the LLM generates case narratives without access to the disease label, which promotes diversity by preventing the model from defaulting to prototypical or template-like disease presentations. Moreover, the first-round generation is conditioned on the primary clinical anchors $\mathcal{P}_{prim}$, ensuring that diversity arises around clinically salient evidence rather than arbitrary variation. In the second round, the disease label is provided to enable targeted correction of obvious inaccuracies, improving factual consistency while retaining most of the diversity introduced in the initial generation. Let $\mathcal{D}_{\text{syn}} = \{(q_i, y_i)\}_{i=1}^{N}$ denote the resulting collection of simulated cases $q_i$ paired with their rare disease labels $y_i$.

Combining the real-world case reports curated from RareArena $\mathcal{D}_{\text{real}}$, we have the initial rare disease case corpus $\mathcal{D}_{\text{case}} = \mathcal{D}_{\text{syn}} \cup \mathcal{D}_{\text{real}}$.

### Model post-training and reasoning enhancement

Starting from the case corpus $\mathcal{D}_{\text{case}}$ defined above, where each case narrative $q_i$ is paired with a rare disease label $y_i$ , we adopted a two-stage post-training pipeline to progressively enhance the diagnostic reasoning ability of the LLM. For each case, the target output consisted of a clinically grounded chain-of-thought reasoning trace $o_i$ together with the corresponding rare disease diagnosis. The overall goal was to train the model to map clinical case descriptions to reliable diagnostic reasoning processes and accurate rare disease predictions.

For each $(q_i, y_i) \in \mathcal{D}_{\text{case}}$, we prompted the advanced reasoning model DeepSeek-R1 to generate a long-form chain-of-thought explanation and a final diagnosis. The generated responses were automatically verified following the RareArena evaluation setting, where a response was considered valid if the ground-truth diagnosis was successfully retrieved with a score of 2 under recall@5. Correct first-pass responses were retained directly. For cases in which the first-pass answer was incorrect, we provided the ground-truth disease label and prompted the model to perform reflection and generate a second complete reasoning attempt. Corrected second-pass reasoning traces were also included in the curated dataset. This process yielded a reasoning corpus $\mathcal{D}_{\text{CoT}} = \{(q_i, o_i)\}_{i=1}^{\text{N}'}$ of size N′=153,836. For the base student model, we selected DeepSeek-R1-Distill-Qwen-32B because it provides a favorable balance among built-in reasoning capability, training efficiency, and practical deployment feasibility. We then fine-tuned the student model on $\mathcal{D}_{\text{CoT}}$ using supervised fine-tuning. This stage encouraged the model to imitate high-quality rare disease diagnostic reasoning traces, thereby equipping it with domain-specific reasoning patterns while preserving the general language capabilities of the base model. The resulting model was denoted as $\mathcal{M}_{\text{SFT}}$.

Although supervised fine-tuning improved diagnostic performance, its effectiveness remained constrained by the quality and coverage of the teacher-generated traces. To further enhance the model's reasoning capability, we applied Direct Preference Optimization (DPO) [17], which optimizes the model using pairwise preference data rather than relying solely on imitation. Starting from $\mathcal{M}_{\text{SFT}}$, we generated multiple reasoning rollouts for selected cases and applied the same automated verification procedure to identify correct and incorrect diagnostic outputs. Reasoning traces leading to correct diagnoses were treated as preferred responses, whereas those leading to incorrect diagnoses were treated as dispreferred responses. In practice, we randomly sampled five cases per disease and performed four rounds of rollout for each case summary to construct preference pairs. To balance diagnostic performance and training cost, we limited the number of preference pairs for each disease to at most three, as pilot experiments on smaller LLMs showed no significant performance gains from including additional pairs. This process produced a preference dataset $\mathcal{D}_{\text{DPO}}$ containing 14,250 triples $(q_i, o_i^+, o_i^-)$, where $o_i^+$ and $o_i^-$ denote the preferred and dispreferred reasoning traces for the same case, respectively. During DPO training, $\mathscr{M}_{\text{SFT}}$ was used as the reference model. The optimization encouraged the current model to assign higher preference to clinically accurate reasoning traces than to weaker or incorrect alternatives. Through this preference-driven contrastive learning process, DPO further refined the model's diagnostic reasoning behavior and improved its ability to generate accurate rare disease predictions beyond what could be achieved through supervised imitation alone.

**Retrospective cohort Study**

**1.1 Participant enrollment and study design**

To evaluate whether RaDaR can generate early differential diagnoses that include the final rare disease diagnosis, we conducted a retrospective cohort study. We retrospectively extracted longitudinal electronic health record (EHR) data for patients with rare diseases, including inpatient admissions, outpatient visits, and emergency department encounters from PUMCH and XHCMU.

Patients were eligible for inclusion if they met the following criteria:

1. Received a confirmed final diagnosis of a rare disease at one of the two participating centers.

2. Lacked both a confirmed diagnosis and a clearly documented suspicion of the final condition at the time of their initial presentation.

We excluded patients if:

1. The final rare disease diagnosis had not yet been confirmed at the time of data extraction.

2. The diagnosis had already been established at an external institution prior to the first visit.

### 1.2 Determination of the Time of Clinical Suspicion

To enable a fair comparison with standard clinical care, we established a physician baseline. Specifically, we defined the physician baseline as the earliest time point at which the final diagnosis was first recorded in the EHR as a clinical suspicion—or could reasonably be inferred from EHR documentation—rather than the time of diagnostic confirmation. The baseline was determined from two sources of evidence. The first is the explicit documentation, which means the specific disease name was explicitly mentioned in the clinical notes or listed in the differential diagnosis. The second is disease-specific testing. Even if the disease name is not explicitly documented, ordering a highly disease-specific test was taken as evidence of clinical suspicion. For example, ordering a *Bartonella henselae* IgG/IgM antibody test for suspected cat-scratch disease. Non–disease-specific investigations, such as whole-genome sequencing or generic pathology for undifferentiated tumors, were not considered sufficient to indicate suspicion, as they do not target a single disease or a narrow set of candidate diagnoses. The longitudinal EHRs were independently reviewed by two reviewers (S.H. and X.J.) to annotate the timing of clinical suspicion, with disagreements adjudicated by a senior rare-disease expert (S.M.).

## 2. Deployment and Evaluation of RaDaR

To simulate real-time clinical progression, we constructed cumulative EMR snapshots by sequentially aggregating data from the first visit through each subsequent encounter. This evaluation sequence continued up to the specific visit where clinical suspicion was first established (the physician baseline). For this final snapshot, clinical notes were strictly truncated immediately prior to the timestamp of the documented suspicion, ensuring the model utilized only information available to the physician before the diagnosis was explicitly considered. At each encounter, RaDaR was queried to generate a differential diagnosis list up to 10 candidates with the prompt listed in Appendix L.

Two reviewers (J.X. and S.H.) independently assessed whether RaDaR's output effectively identified the final confirmed rare disease. The generated differential lists were scored using a 5-point ordinal scale adapted from prior studies (details in Supplementary Table 1). Scores were dichotomized into negative results (scores 0–1; final diagnosis absent or not meaningfully identified) and positive results (scores 2–4; final diagnosis present or strongly implied). For each patient, we identified the RaDaR suspicion timing, defined as the earliest visit number in the sequence at which RaDaR achieved a positive result.

The top1, top3, top5, and full list recall were evaluated. Top-5 recall is reported in the main Results, while the remaining metrics are presented in the Extended Data. Model performance was summarized using two metrics. First, we calculated the visit-normalized diagnostic advancement score (VNDAS), defined as:

$$\mathrm{VNDAS} = 1 - V_{\mathrm{model}} / (V_{\mathrm{all}} + 1)$$

where Vmodel denotes the earliest visit at which the model generated the correct diagnosis, and $V_{all}$ denotes the total number of visits for that patient. A higher VNDAS indicates that the model reached the correct diagnosis earlier in the clinical trajectory. Second, we calculated the proportion of cases in which RaDaR's diagnostic suspicion occurred earlier than, at the same visit as, or later than the physician's clinical suspicion.

**3. Clinical information analysis**

To analyze the flow of clinical information across diagnostic trajectories, we manually reviewed each patient's longitudinal electronic medical records and annotated visit-level changes in clinical information. Clinical information was classified into four categories: symptoms and signs, laboratory tests, imaging and procedures including pathology, and treatments. For the first visit, all recorded information was labeled as newly added. For each subsequent visit, information was classified as newly added, updated, or unchanged.

To explore which types of clinical information contributed to transitions from incorrect to correct diagnoses, we fitted logistic regression models restricted to transitions in which the model prediction was incorrect at the previous visit. The binary outcome was whether the model prediction became correct at the current visit. Predictor variables included the numbers of newly added or updated symptoms, laboratory findings, imaging/procedural findings, and treatments at the current visit. Odds ratios and P values were reported.

**Randomized physician-assistance trial**

The study was approved by the Tsinghua University Ethics Committee (approval number: THU-01-2025-1070 on November 18, 2025) and was prospectively registered in the Chinese Clinical Trial Registry (ChiCTR2500115619). All participants provided written informed consent prior to enrollment. The trial followed the Consolidated Standards of Reporting Trials (CONSORT) guidelines.

1. Physicians Enrollment

Physicians were recruited through online links disseminated via social media platforms between January 1 and January 5, 2026. Interested participants completed an enrollment survey collecting demographic and professional information, including hospital affiliation, clinical department, years of practice, and attitudes toward the use of LLMs in rare disease diagnosis. Eligible participants were practicing physicians from pediatrics, internal medicine, or neurology who were actively involved in clinical practice at the time of enrollment. After completion of the enrollment questionnaire and confirmation of eligibility, participants were randomized and formally enrolled in the trial. We calculated that 35 participants per group were needed to detect a 16% between-group difference in diagnostic performance with 80% power and a two-sided significance level of 0.05, assuming a standard deviation of 24% based on a pilot study. To allow for an anticipated dropout rate of approximately 15%, we increased the target enrollment to 42 participants per group.

2. Study design

This study adopted a two-arm randomized physician-assistance design. Physicians were stratified by clinical department (pediatrics, internal medicine, or neurology) and by years of clinical experience (>5 years or 0–5 years), and were then randomly assigned to either a control arm or a RaDaR-assisted arm using permuted blocks of size two. To balance case difficulty between arms, the two physicians within the same block were assigned the same case set. The allocation sequence was generated using a prespecified Python script and implemented by an independent researcher (S.H.) who was not involved in participant recruitment, enrollment, or outcome assessment. Physicians in the control arm had access to general internet resources but were explicitly instructed not to use any LLMs. Physicians in the RaDaR-assisted arm had access to both general internet resources and RaDaR and were allowed to use RaDaR freely.

### 3. Clinical cases

To closely approximate real-world clinical scenarios, rare disease cases were selected from an in-house dataset at Peking Union Medical College Hospital (PUMCH). All cases had confirmed rare disease diagnoses. Clinical narratives were truncated immediately before the diagnosis was explicitly stated or before definitive diagnostic tests, such as genetic or pathogen testing, were introduced. In total, 50 cases were included, comprising 20 cases from internal medicine, 20 from pediatrics, and 10 from neurology. These cases were organized into four case sets for internal medicine, four for pediatrics, and two for neurology, with each set containing 5 cases.

### 4. Study procedure

Each physician was assigned one department-specific case set containing five cases. For each case, physicians first provided a diagnosis based solely on their own knowledge and experience, serving as the baseline assessment. They then reassessed the same cases under their assigned condition, using either internet resources alone or internet resources plus RaDaR. All diagnostic responses were submitted through a web-based platform, which automatically recorded the time spent on each case. After completing each case, physicians rated the perceived helpfulness of the assigned assistance strategy and their diagnostic confidence on a 5-point Likert scale. Physicians who did not complete the diagnostic tasks were excluded from the final analysis.

### 5. Study outcome

The primary outcome was physician-level diagnostic accuracy, defined as the proportion of the five assigned cases for which the physician provided the correct final diagnosis. Two reviewers (J.Z. and X.J.) were blinded to intervention assignment and participant identifiers and independently evaluated all diagnostic responses, and any discrepancies were resolved by a senior physician (M.S.). Scores of “2 (closely related),” “3 (exact match),” and “4 (more specific and accurate)” were considered correct. The detailed scoring criteria are provided in the Supplementary Table 1.

Secondary outcomes included diagnostic time per case and subjective helpfulness of the assigned assistance strategy. In an exploratory safety-related analysis, we also assessed confidence calibration error, defined as the mean absolute difference across the five cases between diagnostic correctness, coded as 0 or 1, and self-reported confidence, after rescaling confidence scores to a 0–1 range.

**Statistical Analysis**

We used unpaired two-sided t-tests to compare diagnostic advancement scores and two-sided Fisher's exact tests to compare diagnostic advancement ratios across disease subgroups. For the randomized physician-assistance trial, primary and secondary outcomes were compared between the RaDaR-assisted and control arms using two-sided unpaired Student's t-tests. To assess whether the effect of RaDaR differed across physician subgroups, we performed interaction analyses using binomial generalized linear models with a logit link. In these models, the number of correctly diagnosed cases out of the five assigned cases was modeled as the outcome, intervention arm was included as the main predictor, and an interaction term between intervention arm and either clinical experience or department was added. Odds ratios and two-sided P values were reported for the interaction models. Pediatrics and physicians with 0–5 years of clinical experience were used as reference categories. A two-sided $P < 0.05$ was considered statistically significant.

REFERENCES

1 Nguengang Wakap, S. *et al.* Estimating cumulative point prevalence of rare diseases: analysis of the Orphanet database. *European Journal of Human Genetics* **28**, 165-173 (2020). https://doi.org/10.1038/s41431-019-0508-0

2 The Lancet Diabetes, E. Rare diseases: individually rare, collectively common. *Lancet Diabetes Endocrinol* **11**, 139 (2023). https://doi.org/10.1016/s2213-8587(23)00042-6

3 Haendel, M. *et al.* How many rare diseases are there? *Nat Rev Drug Discov* **19**, 77-78 (2020). https://doi.org/10.1038/d41573-019-00180-y

4 Benito-Lozano, J., López-Villalba, B., Arias-Merino, G., Posada de la Paz, M. & Alonso-Ferreira, V. Diagnostic delay in rare diseases: data from the Spanish rare diseases patient registry. *Orphanet J Rare Dis* **17**, 418 (2022). https://doi.org/10.1186/s13023-022-02530-3

5 Taruscio, D. & Gahl, W. A. Rare diseases: challenges and opportunities for research and public health. *Nature Reviews Disease Primers* **10**, 13 (2024). https://doi.org/10.1038/s41572-024-00505-1

6 Glaubitz, R. *et al.* The cost of the diagnostic odyssey of patients with suspected rare diseases. *Orphanet J Rare Dis* **20**, 222 (2025). https://doi.org/10.1186/s13023-025-03751-y

7 Grier, J., Hirano, M., Karaa, A., Shepard, E. & Thompson, J. L. P. Diagnostic odyssey of patients with mitochondrial disease: Results of a survey. *Neurol Genet* **4**, e230 (2018). https://doi.org/10.1212/nxg.0000000000000230

8 Thompson, J. L. P. *et al.* The evolution of the mitochondrial disease diagnostic odyssey. *Orphanet J Rare Dis* **18**, 157 (2023). https://doi.org/10.1186/s13023-023-02754-x

9 Daich Varela, M., Schlottmann, P., Luna Pinto, J. & Michaelides, M. Diagnostic Odyssey of More than 1000 Patients with Inherited Retinal Diseases. *Ophthalmology* **131**, 251-253 (2024). https://doi.org/10.1016/j.ophtha.2023.10.017

10 Zurynski, Y. *et al.* Australian children living with rare diseases: experiences of diagnosis and perceived consequences of diagnostic delays. *Orphanet J Rare Dis* **12**, 68 (2017). https://doi.org/10.1186/s13023-017-0622-4

11 Isono, M., Kokado, M. & Kato, K. Why does it take so long for rare disease patients to get an accurate diagnosis?-A qualitative investigation of patient experiences of hereditary angioedema. *PLoS One* **17**, e0265847 (2022). https://doi.org/10.1371/journal.pone.0265847

12 Miteva, T., Jordanova, R., Iskrov, G. & Stefanov, R. General knowledge and awareness on rare diseases among general practitioners in Bulgaria. *Georgian Med News*, 16-19 (2011).

13 Mao, X. *et al.* A phenotype-based AI pipeline outperforms human experts in differentially diagnosing rare diseases using EHRs. *npj Digital Medicine* **8**, 68 (2025). https://doi.org/10.1038/s41746-025-01452-1

14 Smedley, D. *et al.* Next-generation diagnostics and disease-gene discovery with the Exomiser. *Nat Protoc* **10**, 2004-2015 (2015). https://doi.org/10.1038/nprot.2015.124

15 Tu, T. *et al.* Towards conversational diagnostic artificial intelligence. *Nature* **642**, 442-450 (2025). https://doi.org/10.1038/s41586-025-08866-7
16 Zhou, S. *et al.* Large language models for disease diagnosis: a scoping review. *npj Artificial Intelligence* **1**, 9 (2025). https://doi.org/10.1038/s44387-025-00011-z
17 Liu, X. *et al.* A generalist medical language model for disease diagnosis assistance. *Nature Medicine* **31**, 932-942 (2025). https://doi.org/10.1038/s41591-024-03416-6
18 Shyr, C. *et al.* Large Language Models for Rare Disease Diagnosis at the Undiagnosed Diseases Network. *JAMA Network Open* **8**, e2528538-e2528538 (2025). https://doi.org/10.1001/jamanetworkopen.2025.28538
19 Chimirri, L. *et al.* Consistent performance of large language models in rare disease diagnosis across ten languages and 4917 cases. *eBioMedicine* **121** (2025). https://doi.org/10.1016/j.ebiom.2025.105957
20 Zhao, W. *et al.* An agentic system for rare disease diagnosis with traceable reasoning. *Nature* (2026). https://doi.org/10.1038/s41586-025-10097-9
21 Zhao, W. *et al.* An Agentic System for Rare Disease Diagnosis with Traceable Reasoning. arXiv:2506.20430 (2025). <https://ui.adsabs.harvard.edu/abs/2025arXiv250620430Z>.
22 Guo, D. *et al.* DeepSeek-R1 incentivizes reasoning in LLMs through reinforcement learning. *Nature* **645**, 633-638 (2025). https://doi.org/10.1038/s41586-025-09422-z
23 Riedemann, L., Labonne, M. & Gilbert, S. The path forward for large language models in medicine is open. *npj Digital Medicine* **7**, 339 (2024). https://doi.org/10.1038/s41746-024-01344-w
24 Nagarajan, R. *et al.* Economics and Equity of Large Language Models: Health Care Perspective. *J Med Internet Res* **26**, e64226 (2024). https://doi.org/10.2196/64226
25 Haichao Chen, Z. Z., Songchi Zhou, Shikai Hu, Jinyuan Wang, Ye Jin, Xianghong Jin, Xiaofei Wang, Weizhi Ma, Yih Chung Tham, Honghan Wu, Bin Sheng, Shuyang Zhang, Sheng Yu, Tien Yin Wong. RareArena: A Comprehensive Benchmark Dataset Unveiling the Potential of Large Language Models in Rare Disease Diagnosis. *Lancet Digital Health* **In Press** (2025).
26 Chen, X. *et al.* RareBench: Can LLMs Serve as Rare Diseases Specialists? , arXiv:2402.06341 (2024). <https://ui.adsabs.harvard.edu/abs/2024arXiv240206341C>.
27 Reese, J. T. *et al.* Systematic benchmarking demonstrates large language models have not reached the diagnostic accuracy of traditional rare-disease decision support tools. *European Journal of Human Genetics* **34**, 498-504 (2026). https://doi.org/10.1038/s41431-026-02054-5
28 OpenAI *et al.* gpt-oss-120b & gpt-oss-20b Model Card. arXiv:2508.10925 (2025). <https://ui.adsabs.harvard.edu/abs/2025arXiv250810925O>.
29 Chen, J. *et al.* HuatuoGPT-o1, Towards Medical Complex Reasoning with LLMs. arXiv:2412.18925 (2024). <https://ui.adsabs.harvard.edu/abs/2024arXiv241218925C>.
30 Huang, X., Wu, J., Liu, H., Tang, X. & Zhou, Y. m1: Unleash the Potential of Test-Time Scaling for Medical Reasoning with Large Language Models. arXiv:2504.00869 (2025). <https://ui.adsabs.harvard.edu/abs/2025arXiv250400869H>.
31 Team, M. *et al.* Baichuan-M2: Scaling Medical Capability with Large Verifier System. arXiv:2509.02208 (2025). <https://ui.adsabs.harvard.edu/abs/2025arXiv250902208M>.

32 Fujiwara, T., Shin, J. M. & Yamaguchi, A. Advances in the development of PubCaseFinder, including the new application programming interface and matching algorithm. *Hum Mutat* **43**, 734-742 (2022). https://doi.org/10.1002/humu.24341
33 Robinson, P. N. *et al.* Interpretable Clinical Genomics with a Likelihood Ratio Paradigm. *Am J Hum Genet* **107**, 403-417 (2020). https://doi.org/10.1016/j.ajhg.2020.06.021
34 Kobayashi Takahashi, Y., Hayakawa, I. & Abe, Y. Diagnostic odyssey of Guillain-Barré syndrome in children. *Brain Dev* **46**, 108-113 (2024). https://doi.org/10.1016/j.braindev.2023.10.004
35 Garcia, B. T. *et al.* Improving automated deep phenotyping through large language models using retrieval-augmented generation. *Genome Med* **17**, 91 (2025). https://doi.org/10.1186/s13073-025-01521-w
36 Yang, J. *et al.* Enhancing phenotype recognition in clinical notes using large language models: PhenoBCBERT and PhenoGPT. *Patterns* **5**, 100887 (2024). https://doi.org/https://doi.org/10.1016/j.patter.2023.100887
37 Savage, C. H. *et al.* Open-Source Large Language Models in Radiology: A Review and Tutorial for Practical Research and Clinical Deployment. *Radiology* **314**, e241073 (2025). https://doi.org/10.1148/radiol.241073
38 Wiest, I. C. *et al.* Privacy-preserving large language models for structured medical information retrieval. *NPJ Digit Med* **7**, 257 (2024). https://doi.org/10.1038/s41746-024-01233-2
39 McIntosh, F., Murina, S., Chen, L., Vargas, H. A. & Becker, A. S. Keeping private patient data off the cloud: A comparison of local LLMs for anonymizing radiology reports. *European Journal of Radiology Artificial Intelligence* **2**, 100020 (2025). https://doi.org/https://doi.org/10.1016/j.ejrai.2025.100020
40 Shea, Y. F., Lee, C. M. Y., Ip, W. C. T., Luk, D. W. A. & Wong, S. S. W. Use of GPT-4 to Analyze Medical Records of Patients With Extensive Investigations and Delayed Diagnosis. *JAMA Netw Open* **6**, e2325000 (2023). https://doi.org/10.1001/jamanetworkopen.2023.25000
41 Mitani, A. A. & Haneuse, S. Small Data Challenges of Studying Rare Diseases. *JAMA Netw Open* **3**, e201965 (2020). https://doi.org/10.1001/jamanetworkopen.2020.1965
42 May, M. Rare-disease researchers pioneer a unique approach to clinical trials. *Nature Medicine* **29**, 1884-1886 (2023). https://doi.org/10.1038/s41591-023-02333-4
43 Reese, J. T. *et al.* Evaluation of the Diagnostic Accuracy of GPT-4 in Five Thousand Rare Disease Cases. *medRxiv*, 2024.2007.2022.24310816 (2024). https://doi.org/10.1101/2024.07.22.24310816

## Figure Legend

### Figure 1. Development and multi-stage validation framework of the RaDaR model.

The top left panel illustrates the model development pipeline, consisting of three stages: phenotype construction, synthetic case augmentation, and reasoning enhancement. The top right panel summarizes the three major steps of model validation. The first is the cross-sectional evaluation, including both public benchmark evaluation and external validation (334 cases). The second is a retrospective cohort study, which assesses model's performance along retrospective rare-disease diagnostic trajectories. The third is a randomized physician-assistance trial, which evaluates whether RaDaR assistance improves physicians' rare-disease diagnostic accuracy. The bottom table presents benchmark details, including the number of cases and diseases, input types, and data accessibility status.

### Figure 2. Diagnostic performance of RaDaR and baseline models on rare disease benchmarks and external validation.

(a) Internal validation on the RareArena hold-out test set for two tasks: Rare Disease Screening (left) and Rare Disease Confirmation (right). (b) External validation performance on real-world patient datasets collected from four independent clinical centers. Bars indicate diagnostic accuracy, defined as top-5 recall; error bars indicate standard errors of the mean across 10 repeated evaluations. (c) Model performance on the two phenotype-input rare disease benchmarks (left, Reese et al.; right, RareBench)

### Figure 3. Evaluation of RaDaR performance along rare-disease diagnostic trajectories.

(a) Schematic illustration of the study design comparing the timing of diagnosis. The timelines depict the clinical journey from symptom onset to documented clinical suspicion. Two distinct phases are identified: the pre-center interval (symptom onset to initial presentation at a participating center) and the within-center interval (initial presentation to clinical suspicion). The timing of the model's correct identification (RaDaR suspicion, red cross) is compared with the initial clinical suspicion documented by physicians (blue cross). (b) Visit-normalized diagnostic advancement scores. Bars indicate mean scores and error bars indicate standard errors of the mean. (c) Distribution of diagnostic timing relative to physician suspicion for each model, categorized as earlier, same visit, or later identification. (d) Potential diagnostic lead time if the final diagnosis had been prioritized at the earliest visit when RaDaR identified it. Horizontal stacked bars show the distribution of lead time measured in months and number of visits. Grey indicates no diagnostic advancement; progressively darker red segments indicate larger gains.

### Figure 4. Evaluation of RaDaR potential in facilitating rare disease diagnosis in a randomized physician-assistance trial.

(a) Study design of the randomized physician-assistance trial. (b) Diagnostic accuracy in the control and RaDaR-assisted groups, defined as the proportion of the five assigned cases for which physicians included the final diagnosis in their differential diagnosis list. (c) Time required to complete the diagnostic task. (d) Self-reported helpfulness score. (e) Diagnostic accuracy stratified by physicians' working experience. (f) Diagnostic accuracy stratified by specialty. Bars indicate mean values and error bars indicate standard errors of the mean. Each dot represents one physician. Diagnostic accuracy was calculated as the proportion of correctly diagnosed cases among five assigned cases. Helpfulness was assessed using a 1–5 rating scale. Statistical comparisons for overall physician-level outcomes were performed using two-sided unpaired Student's t-tests. Interaction analyses using binomial generalized linear models were performed to determine whether the benefit of RaDaR varied across physician subgroups. n.s., not significant; ****$P < 0.0001$.

### Figure 5. Synthetic-data ablation and scaling analyses.

(a) Performance comparison between the baseline model and the synthetic-data-augmented model across four frequency bins. (b) The absolute performance gain (in percentage points) attributed to synthetic augmentation. (c) Overall performance comparison of the base model, the model fine-tuned on real-world data only, and the model fine-tuned on synthetic data only. (d) Performance of the model fine-tuned on synthetic data only, evaluated on synthesizable and non-synthesizable diseases. (e) Synthetic-data scaling trend. The log-log scale plot illustrates the monotonic trend between diagnostic performance and the number of synthetic samples generated per disease (ranging from 1 to 30).

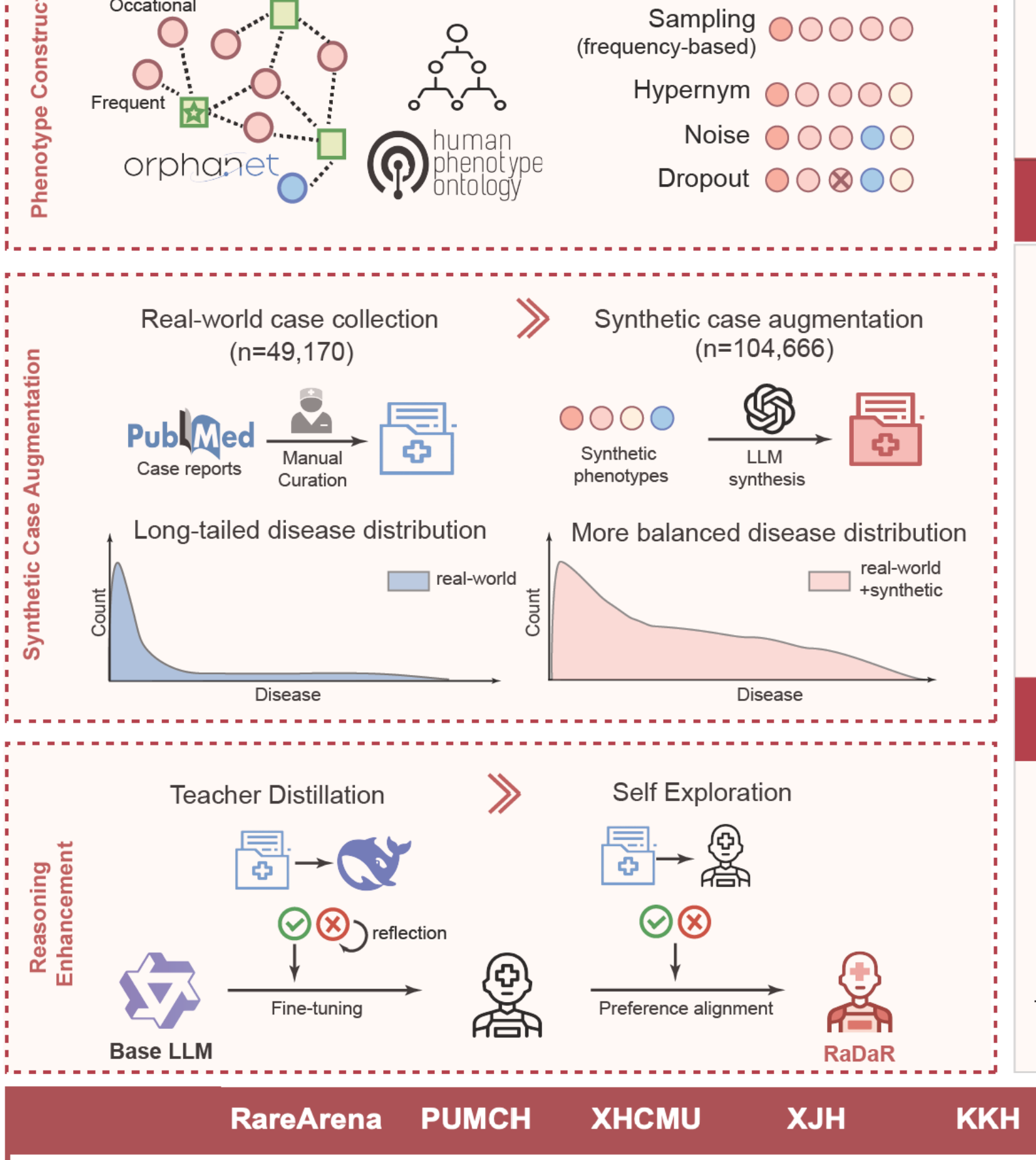

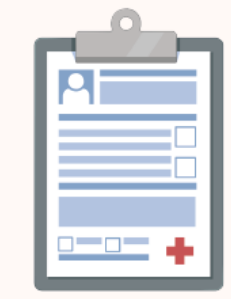
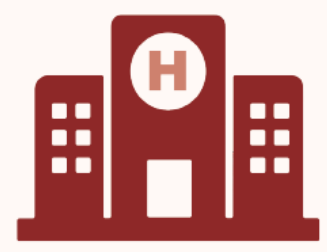

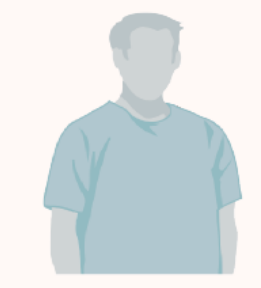
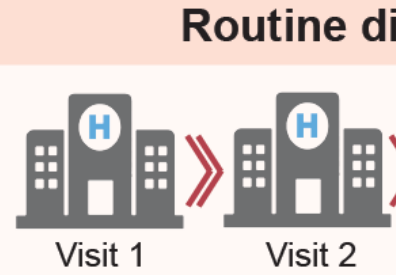
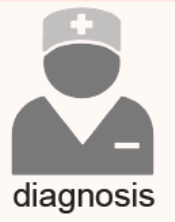
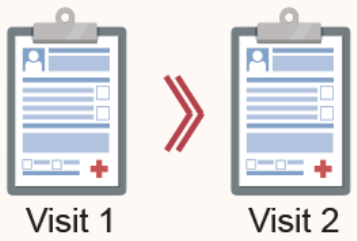
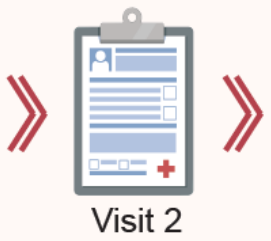
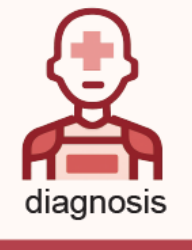

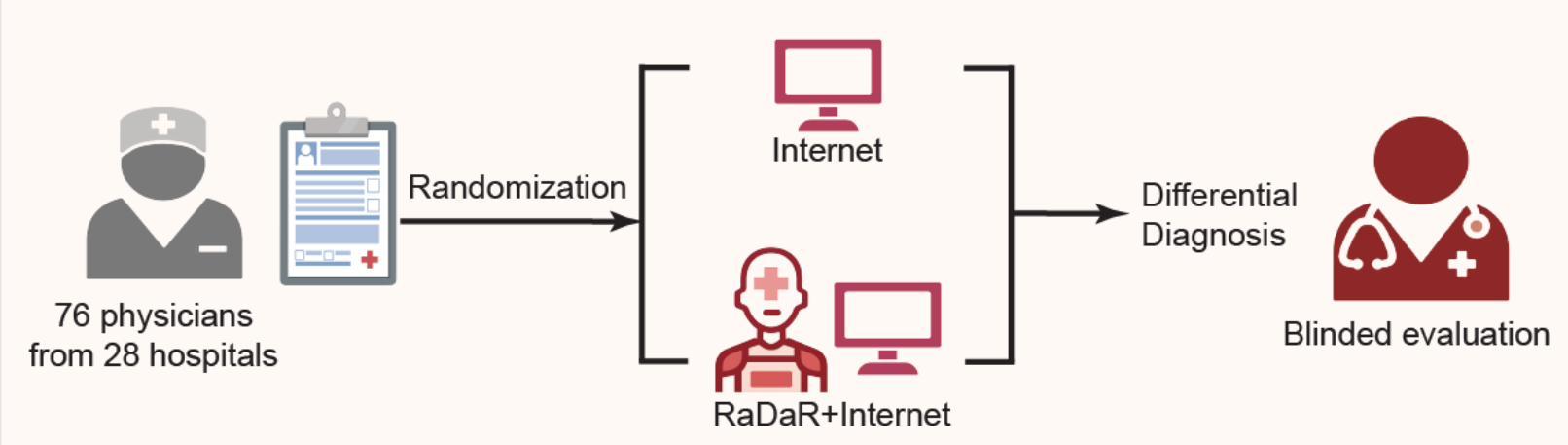

| | RareArena | PUMCH | XHCMU | XJH | KKH | RareBench (MME) | RareBench (HMS) | RareBench (LIRICAL) | RareBench (RAMEDIS) | RareBench (PUMCH) | Reese et al. |
|---|---|---|---|---|---|---|---|---|---|---|---|
| **Cases** | 12938 | 211 | 35 | 27 | 61 | 40 | 88 | 370 | 624 | 75 | 5627 |
| **Rare Diseases** | 3737 | 152 | 18 | 10 | 42 | 17 | 46 | 252 | 74 | 16 | 378 |
| **Type** | Case report | EHR | EHR | EHR | EHR | Phenotype | Phenotype | Phenotype | Phenotype | Phenotype | Phenotype |
| **Public** | ✓ | ✗ | ✗ | ✗ | ✗ | ✓ | ✓ | ✓ | ✓ | ✓ | ✓ |

Free-text Input

a

Internal Validation

Hypernym
Exact match

Model Size(B)

Rare disease confirmation

HuatuoGPT-o1 | 72B
m1 | 32B
Baichuan-M2 | 32B
gpt-oss | 120B
DeepSeek-R1 | 671B
RaDaR | 32B

Medical domain
General

15 25 35 45
Performance (%)

40 50 60 70 80
Performance (%)

b

External Validation

PUMCH XHCMU XJH KKH

HuatuoGPT-o1
m1
Baichuan-M2
gpt-oss
DeepSeek-R1
RaDaR

n = 334 from 4 centers

40 50 60 70 80
60 65 70 75 80
60 70 80 90
10 20 30 40 50 60

Performance (%)

Phenotype Input

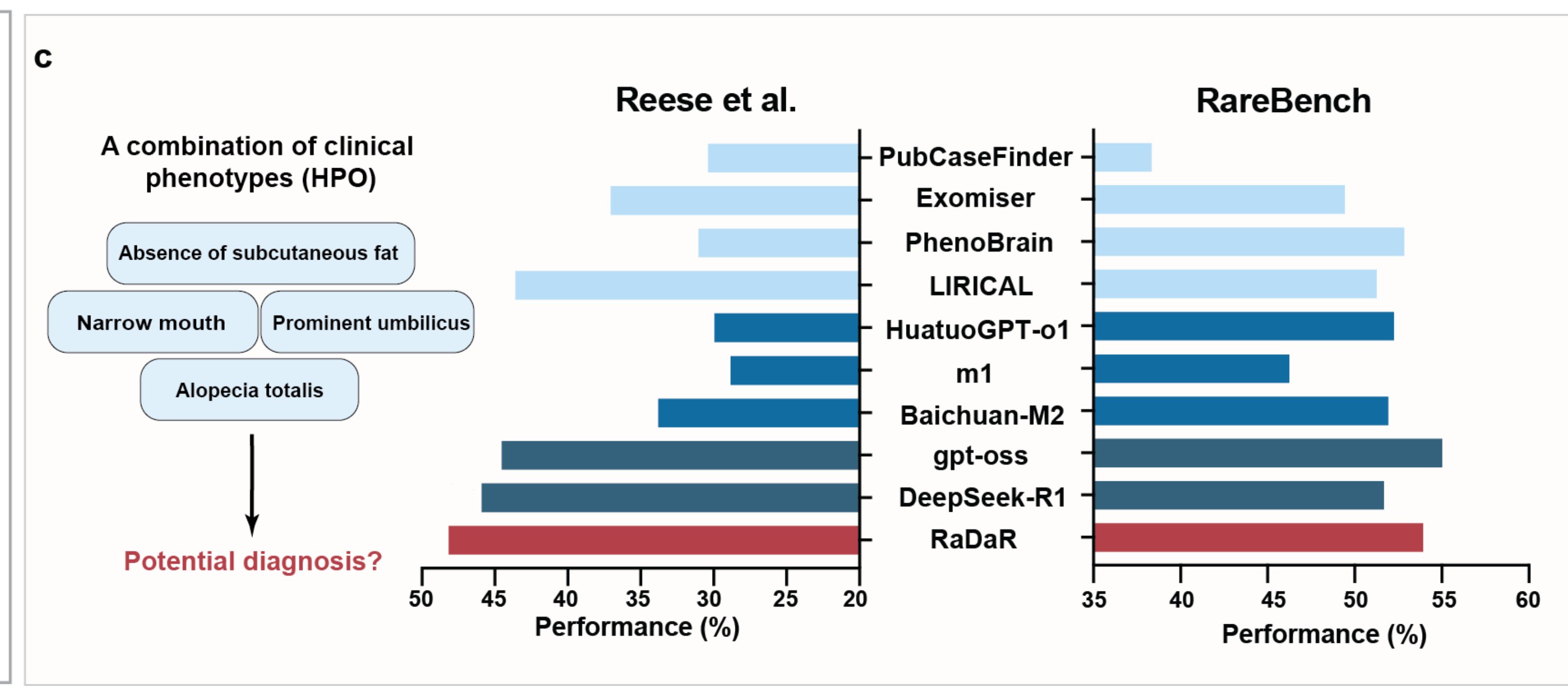

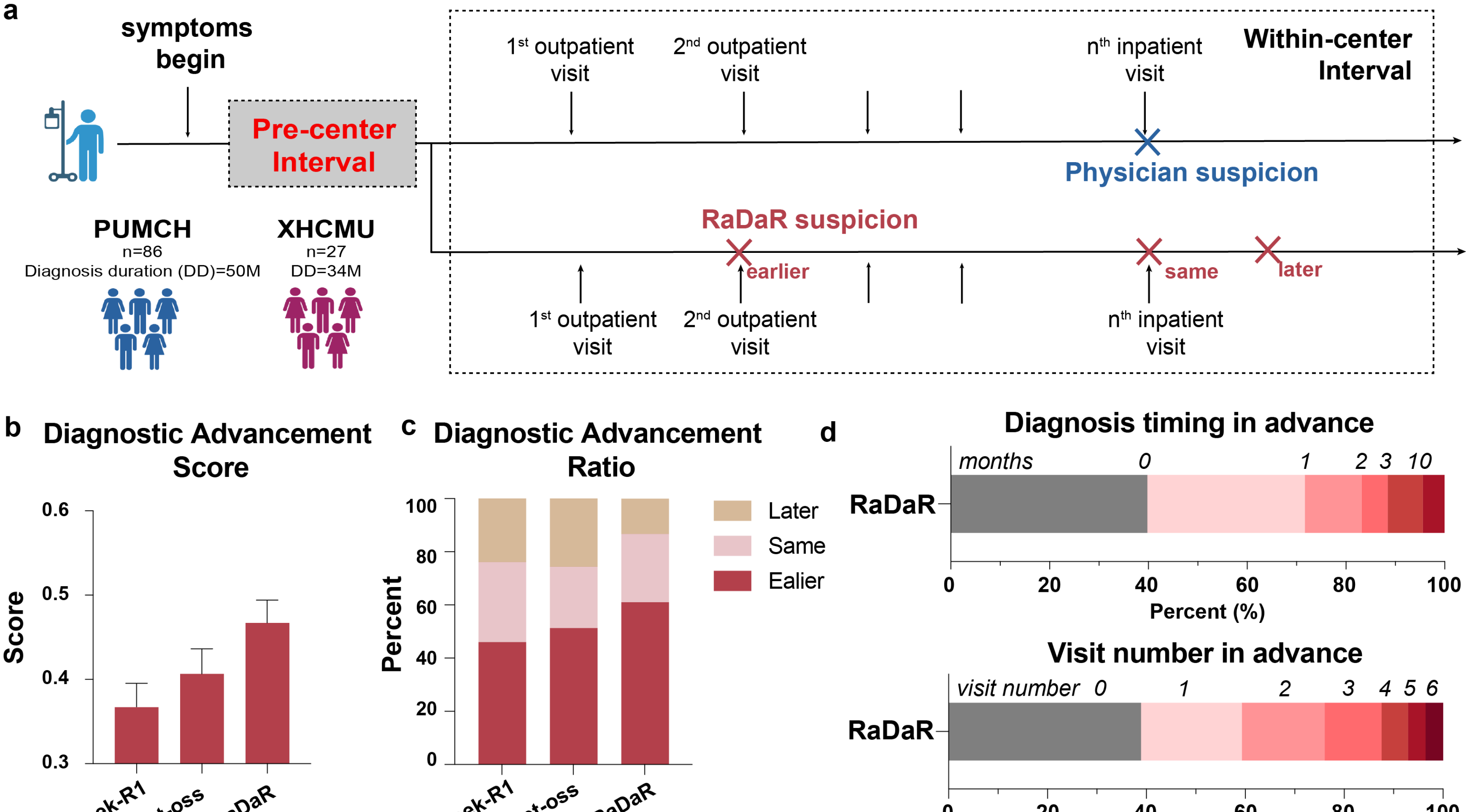
a
symptoms begin
Pre-center Interval
PUMCH
n=86
Diagnosis duration (DD)=50M
XHCMU
n=27
DD=34M
1st outpatient visit
2nd outpatient visit
nth inpatient visit
Within-center Interval
Physician suspicion
RaDaR suspicion
earlier
same
later
b
Diagnostic Advancement Score
Score
DeepSeek-R1
gpt-oss
RaDaR
c
Diagnostic Advancement Ratio
Percent
Later
Same
Ealier
d
Diagnosis timing in advance
months
RaDaR
Percent (%)
Visit number in advance
visit number

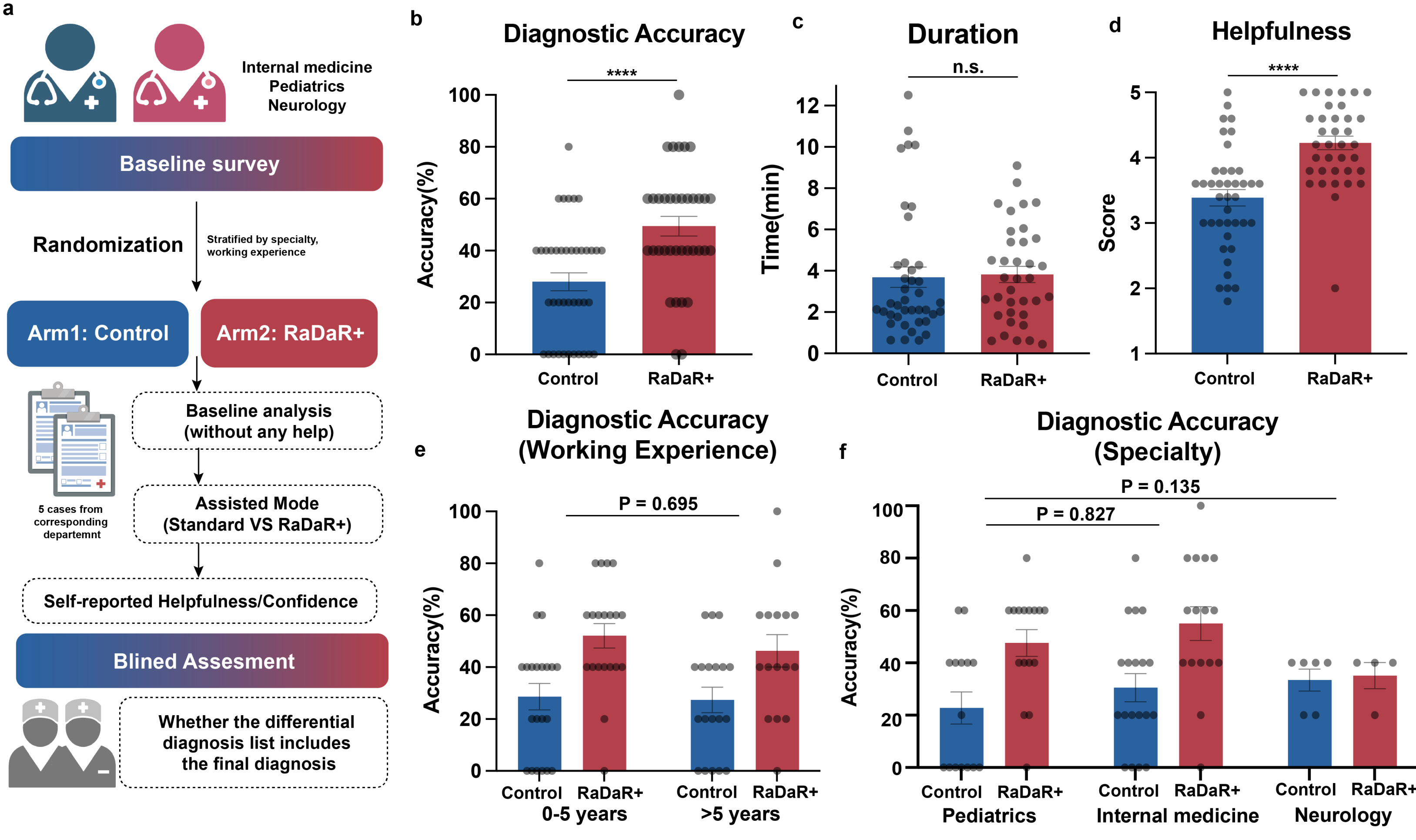
a
Internal medicine
Pediatrics
Neurology
Baseline survey
Randomization
Stratified by specialty, working experience
Arm1: Control
Arm2: RaDaR+
Baseline analysis (without any help)
5 cases from corresponding departemnt
Assisted Mode (Standard VS RaDaR+)
Self-reported Helpfulness/Confidence
Blined Assesment
Whether the differential diagnosis list includes the final diagnosis
b
Diagnostic Accuracy
****
Accuracy(%)
Control
RaDaR+
c
Duration
n.s.
Time(min)
Control
RaDaR+
d
Helpfulness
****
Score
Control
RaDaR+
e
Diagnostic Accuracy (Working Experience)
P = 0.695
Accuracy(%)
Control
RaDaR+
0-5 years
>5 years
f
Diagnostic Accuracy (Specialty)
P = 0.135
P = 0.827
Accuracy(%)
Pediatrics
Internal medicine
Neurology

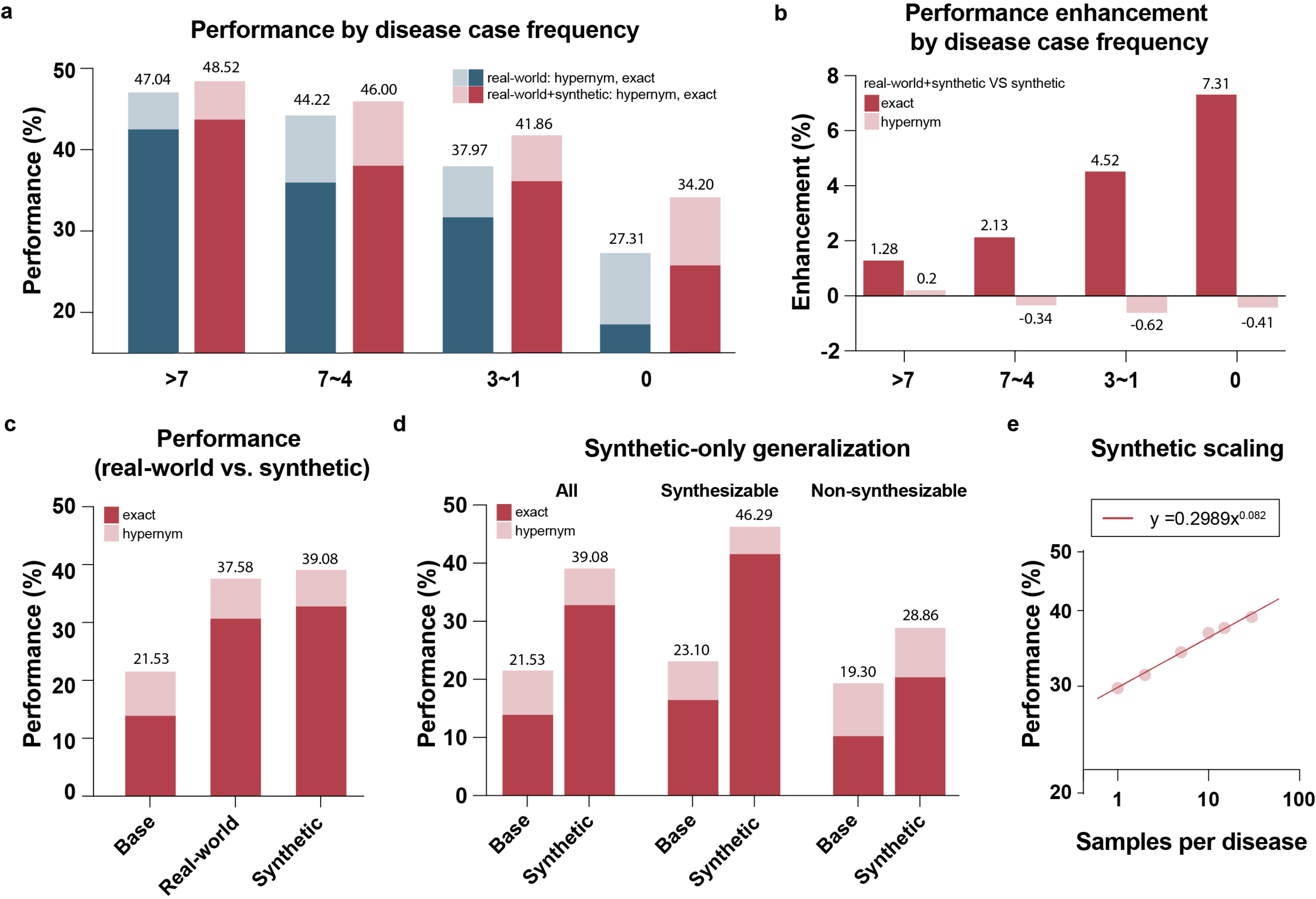
a
Performance by disease case frequency
real-world: hypernym, exact
real-world+synthetic: hypernym, exact
Performance (%)
47.04
48.52
44.22
46.00
37.97
41.86
27.31
34.20
>7
7~4
3~1
0
b
Performance enhancement
by disease case frequency
real-world+synthetic VS synthetic
exact
hypernym
Enhancement (%)
1.28
0.2
2.13
-0.34
4.52
-0.62
7.31
-0.41
c
Performance
(real-world vs. synthetic)
21.53
37.58
39.08
Base
Real-world
Synthetic
d
Synthetic-only generalization
All
Synthesizable
Non-synthesizable
21.53
39.08
23.10
46.29
19.30
28.86
e
Synthetic scaling
y =0.2989x^0.082
Samples per disease